\documentclass{article}
\usepackage[utf8]{inputenc}

\usepackage{microtype}
\usepackage{graphicx}
\usepackage{subfigure}
\usepackage{booktabs} 
\usepackage[most]{tcolorbox}
\usepackage{enumitem}

\usepackage{hyperref}

\usepackage[accepted]{icml2025_arxiv}

\usepackage{amsmath}
\usepackage{amssymb}
\usepackage{mathtools}
\usepackage{amsthm}
\usepackage{xcolor}

\theoremstyle{plain}
\newtheorem{theorem}{Theorem}[section]

\theoremstyle{definition}
\newtheorem{definition}[theorem]{Definition}

\theoremstyle{remark}

\definecolor{Burgundy}{RGB}{144,0,32}

\usepackage[textsize=tiny]{todonotes}

\usepackage[nameinlink,capitalise,noabbrev]{cleveref}
\crefformat{equation}{(#2#1#3)}  

\icmltitlerunning{The Danger of Overthinking: Examining the Reasoning-Action Dilemma in Agentic Tasks}

\definecolor{CiteColor}{RGB}{30,76,132}
\definecolor{LinkColor}{RGB}{0,128,0}
\hypersetup{colorlinks    = true,
            final,
            citecolor=CiteColor,
            linkcolor=LinkColor,
            urlcolor=CiteColor,
}

\begin{document}

\twocolumn[
\icmltitle{The Danger of Overthinking: Examining the Reasoning-Action Dilemma in
Agentic Tasks}



\icmlsetsymbol{equal}{*}

\begin{icmlauthorlist}
\icmlauthor{Alejandro Cuadron}{ucb,ethz}
\icmlauthor{Dacheng Li}{ucb}
\icmlauthor{Wenjie Ma}{ucb}
\icmlauthor{Xingyao Wang}{uiuc}
\icmlauthor{Yichuan Wang}{ucb}
\icmlauthor{Siyuan Zhuang}{ucb}
\icmlauthor{Shu Liu}{ucb}
\icmlauthor{Luis Gaspar Schroeder}{ucb}
\icmlauthor{Tian Xia}{ucb}
\icmlauthor{Huanzhi Mao}{ucb}
\icmlauthor{Nicholas Thumiger}{ethz}
\icmlauthor{Aditya Desai}{ucb}
\icmlauthor{Ion Stoica}{ucb}
\icmlauthor{Ana Klimovic}{ethz}
\icmlauthor{Graham Neubig}{cmu}
\icmlauthor{Joseph E. Gonzalez}{ucb}
\end{icmlauthorlist}

\icmlaffiliation{ucb}{Department of EECS, University of California, Berkeley, USA}
\icmlaffiliation{ethz}{Department of Computer Science, ETH, Zurich, Switzerland}
\icmlaffiliation{uiuc}{Department of Computer Science, University of Illinois Urbana-Champaign, USA}
\icmlaffiliation{cmu}{Department of Computer Science, Carnegie Mellon University, USA}

\icmlcorrespondingauthor{Alejandro Cuadron}{acuadron@berkeley.edu}

\icmlkeywords{Machine Learning, ICML, LLM, reasoning, O1, SWE-Bench, multi-turn environment}

\vskip 0.3in
]



\printAffiliationsAndNotice{}  

\begin{abstract}
Large Reasoning Models (LRMs) represent a breakthrough in AI problem-solving capabilities, but their effectiveness in interactive environments can be limited. This paper introduces and analyzes \textbf{overthinking} in LRMs—a phenomenon where models favor extended internal reasoning chains over environmental interaction. Through experiments on software engineering tasks using SWE Bench Verified, we observe three recurring patterns: \textit{Analysis Paralysis, Rogue Actions, and Premature Disengagement}. We propose a framework to study these behaviors, which correlates with human expert assessments, and analyze \textbf{4018 trajectories}. We observe that higher overthinking scores correlate with decreased performance, with reasoning models exhibiting stronger tendencies toward overthinking compared to non-reasoning models. Our analysis reveals that simple efforts to mitigate overthinking in agentic environments — such as selecting the solution with the lower overthinking score — can \textbf{improve model performance by almost 30\% while reducing computational costs by 43\%}. These results suggest that mitigating overthinking has strong practical implications. We suggest that by leveraging native function-calling capabilities and selective reinforcement learning overthinking tendencies could be mitigated. We also open-source our evaluation framework and dataset to facilitate research in this direction at \url{https://github.com/AlexCuadron/Overthinking}.

\end{abstract}

\section{Introduction}\label{introduction}

\begin{figure}[t]
    \centering
    \includegraphics[width=1\linewidth,trim={0cm 0cm 0cm 0cm},clip]{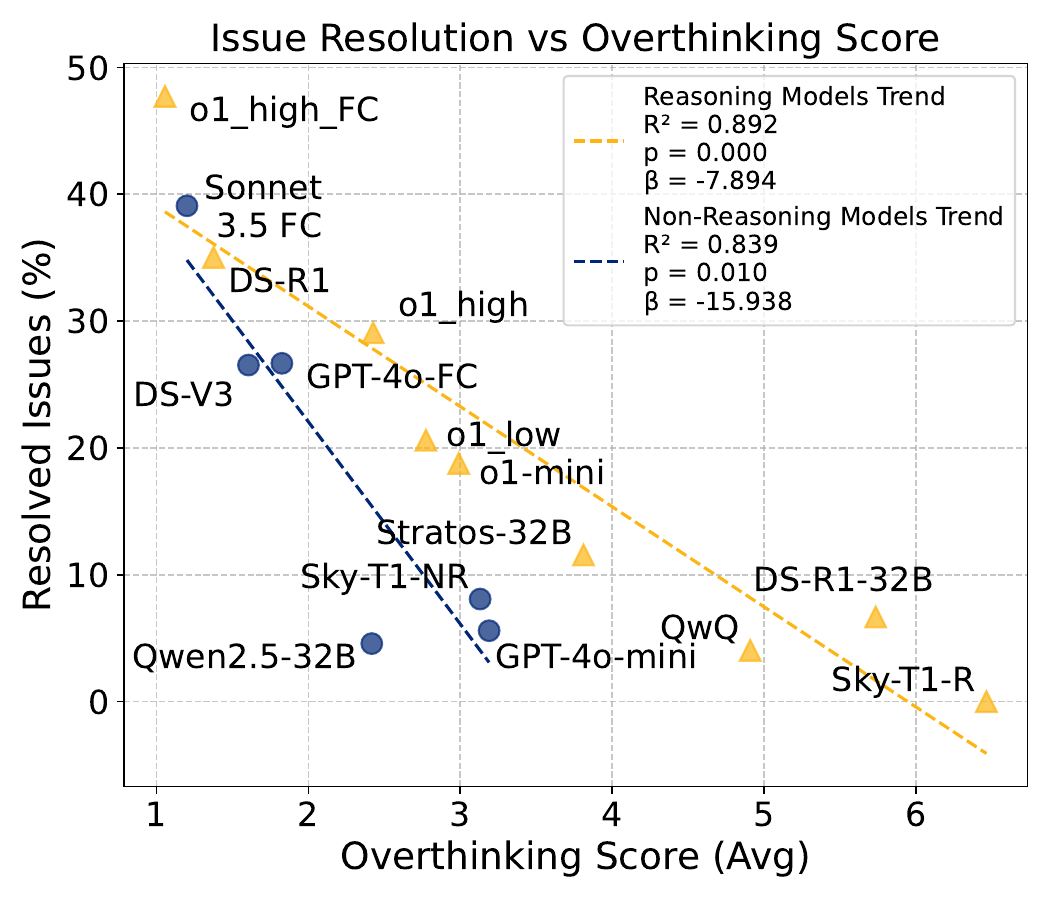}
    \vspace{-25pt}
\caption{
Higher overthinking scores (tendency to favor internal reasoning over environmental feedback) correlate with lower issue resolution rates across all models. Reasoning models exhibit consistently higher overthinking tendencies, suggesting that excessive reliance on internal simulation impairs task performance. Model nomenclature: "FC" indicates native function calling capability, "DS" represents DeepSeek models, and suffixes o1\_high and o1\_low denote models with reasoning effort set to high and low respectively.
}
    \label{fig:figure1}
\end{figure}

\begin{figure}[t]
    \centering
    \includegraphics[width=1\linewidth,trim={0cm 0cm 0cm 0cm},clip]{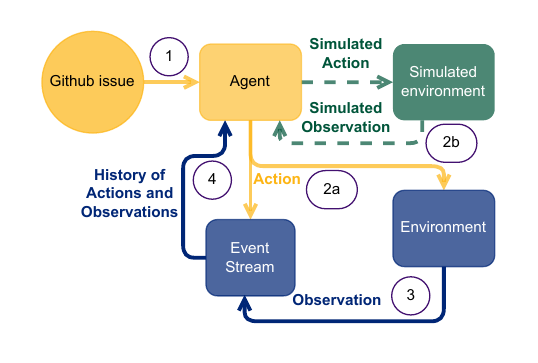}
\caption{OpenHands Execution Pipeline. 1) The system initializes by presenting the agent with the primary issue and previous action history. 2) The agent reaches a decision point -- 2a) Direct action formulation and execution, or 2b) Internal simulation of potential actions and outcomes, potentially leading to \textbf{overthinking}. 3) The chosen action is executed, generating environmental feedback which updates the event stream. This cycle continues until task completion.}
    \label{fig:figure3}
\end{figure}

\begin{figure}[t]
    \centering
    \includegraphics[width=1\linewidth,trim={0cm 0cm 0cm 0cm},clip]{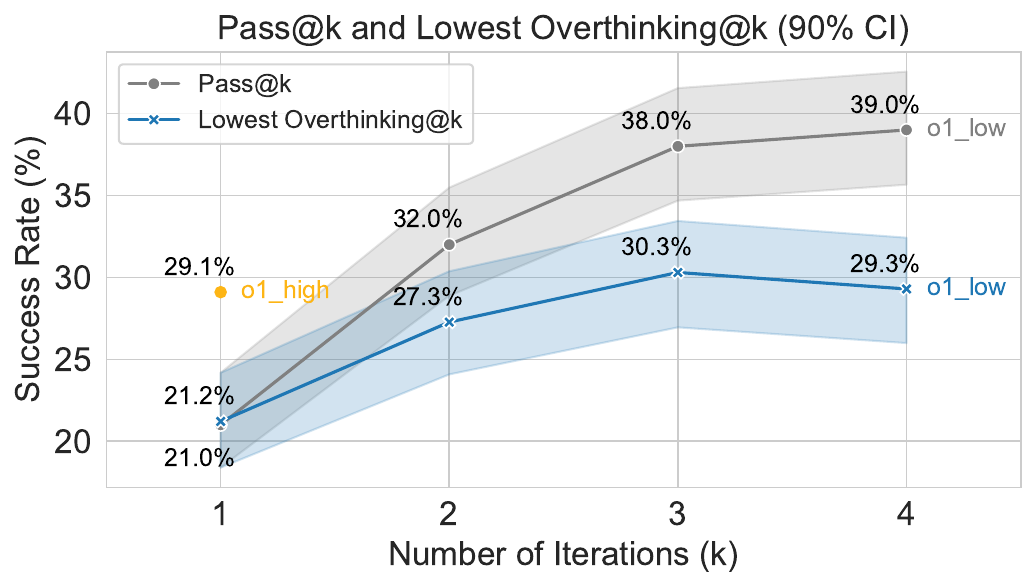}
    \vspace{-25pt}
    \caption{
Performance comparison of Pass@k and Lowest Overthinking@k on SWE-bench Verified. Pass@k represents the success rate when considering k solutions, while Lowest Overthinking@k shows the success rate when selecting the solution with minimal overthinking from k samples. Using k=2 samples with low reasoning effort, we achieve a 27.3\% success rate while reducing computational costs by 43\% compared to high reasoning configurations. Increasing to k=3 further improves performance to 30.3\% surpassing the high configuration using 15\% less computational costs. The confidence intervals (CI) were computed using Wilson score \cite{wallis2013binomial}.}
    \label{fig:figure2}
\end{figure}

Large Reasoning Models (LRMs) \cite{ guan2025rstarmathsmallllmsmaster, xu2025largereasoningmodelssurvey}, such as OpenAI’s
o1 \cite{openai_o1_system_card_2024}, Alibaba’s QwQ \cite{qwq-32b-preview}, or Deepseek's R1 \cite{deepseekai2025deepseekr1incentivizingreasoningcapability} represent a breakthrough in large language models (LLMs). These advanced systems have fundamentally redefined AI’s problem-solving capabilities across various domains \cite{besta2025reasoning}. In particular, LRM's self-correction abilities enable them to achieve impressive scores in several benchmarks, such as AIME 2024 \cite{aops_aime_2024}, MMLU \cite{hendrycks2021measuringmassivemultitasklanguage}, or GPQA-Diamond \cite{rein2023gpqagraduatelevelgoogleproofqa} among others \cite{deepseekai2025deepseekr1incentivizingreasoningcapability, openai_o1_system_card_2024, openai_o1_mini, qwq-32b-preview, guan2025rstarmathsmallllmsmaster}.

Despite extensive analysis of LRMs in non-agentic environments, there remains a critical gap in understanding how LRMs perform in agentic environments \cite{smeyatsky2024agentic}, where models must simultaneously gather, retain, and act upon new information to complete their tasks \cite{zhang2024agenticinformationretrieval,yang2024sweagentagentcomputerinterfacesenable}. In this context, LRMs face a fundamental challenge: models must choose between engaging directly with their environment or relying on internal reasoning about potential actions and their hypothetical consequences, a challenge we define as the \textbf{\textit{Reasoning-Action Dilemma}}.

In this work, we present the first comprehensive empirical study of LRMs in agentic tasks at balancing the Reasoning-Action Dilemma, using real-world software engineering tasks as our experimental framework \cite{jimenez2024swebenchlanguagemodelsresolve, yang2024sweagentagentcomputerinterfacesenable}. We employ \emph{SWE-bench Verified} \cite{jimenez2024swebenchlanguagemodelsresolve, swebench_verified} as our benchmark, using the \emph{CodeAct} agent scaffolding \cite{wang2024executablecodeactionselicit} within the \emph{OpenHands} framework \cite{wang2024openhandsopenplatformai}. This setup creates a controlled environment where models must balance information gathering with reasoning chains while maintaining context across multiple interactions as illustrated in \autoref{fig:figure3}. A proper balance becomes critical as too much reliance on internal reasoning chains might lead to false assumptions about the environment.

We observe that LRMs exhibit a consistent pattern of favoring internal simulation over environmental interaction in the Reasoning-Action Dilemma, spending increasing amounts of time constructing elaborate chains of predicted actions rather than adapting to actual system responses, a phenomenon we define as \textbf{overthinking}.

To quantify \textbf{overthinking}, we develop and validate a systematic evaluation framework using LLM-as-a-judge \cite{zheng2023judgingllmasajudgemtbenchchatbot} that identifies three key patterns: Analysis Paralysis, Rogue Actions, and Premature Disengagement (\cref{fig:manifestations}). Our scoring system strongly correlates with human expert assessments (\cref{fig:human_eval}), confirming its reliability in measuring a model's tendency to favor internal simulation over environmental interaction. We applied this framework to analyze \textbf{4018 trajectories}, creating a comprehensive open-source dataset to advance research in balancing reasoning and action in agentic environments.

Statistical analysis reveals two distinct patterns in overthinking behavior. First, regression analysis demonstrates a significant negative correlation between overthinking and issue resolution rates for both reasoning and non-reasoning models (\cref{fig:figure1}), with the latter showing a steeper decline in performance as overthinking increases. Second, a direct comparison reveals that reasoning models consistently exhibit higher overthinking scores—nearly three times higher than non-reasoning models—with this difference being statistically significant as shown 
 afterward in \cref{tab:overthinking_scores}. These patterns suggest that while all models are susceptible to overthinking, reasoning models are particularly prone to this behavior.

Addressing overthinking yields substantial practical benefits. Running o1 with high reasoning effort achieves 29.1\% issue resolution but costs \$1,400, while the low reasoning variant reaches 21.0\% at 3.5$\times$ lower cost (\$400). Instead of using the expensive high-reasoning configuration, we found that generating two solutions with low reasoning effort (\$800 total) and selecting the one with a lower overthinking score achieves 27.3\% resolution rate (\cref{fig:figure2}). This simple strategy nearly matches the performance of high-reasoning configurations while reducing computational costs by 43\%, demonstrating that overthinking mitigation can dramatically improve the efficiency of LRMs in real-world applications.

Additionally, we suggest two potential approaches to mitigate overthinking in LRMs in agentic environments: native function-calling capabilities and selective reinforcement learning. Both approaches could significantly reduce overthinking while improving model performance, with function-calling models showing particularly promising results (\cref{fixoverthinking}). To facilitate further research into these solutions, we release our evaluation framework and dataset, enabling the broader research community to build upon these findings across different environments and architectures.

\section{Background and Related Work}
In this section, we explore Large Reasoning Models (LRMs) and agentic environments,
where LRMs must balance sophisticated reasoning with practical actions. We examine how these models navigate environments requiring deep analytical thinking and concrete interactions. This leads to a fundamental dilemma between reasoning depth and action—a tension we will explore in \cref{section3}.

\subsection{Large Reasoning Models and Agentic Environments}
LRMs, defined as language models optimized through process reward models and test-time compute scaling \cite{xu2025largereasoningmodelssurvey}, represent an evolution beyond traditional LLMs through their focus on reliable step-by-step reasoning \cite{deepseekai2025deepseekr1incentivizingreasoningcapability,openai_o1}. These models achieve unprecedented performance through extended chain-of-thought reasoning \cite{wei2023chainofthoughtpromptingelicitsreasoning} and rigorous self-verification \cite{madaan2023selfrefineiterativerefinementselffeedback}. However, their extended focus on internal reasoning depth raises important questions about performance in interactive environments.

\begin{figure*}[t]
    \centering
    \includegraphics[width=1\textwidth,trim={0cm 4cm 0cm 1cm},clip]{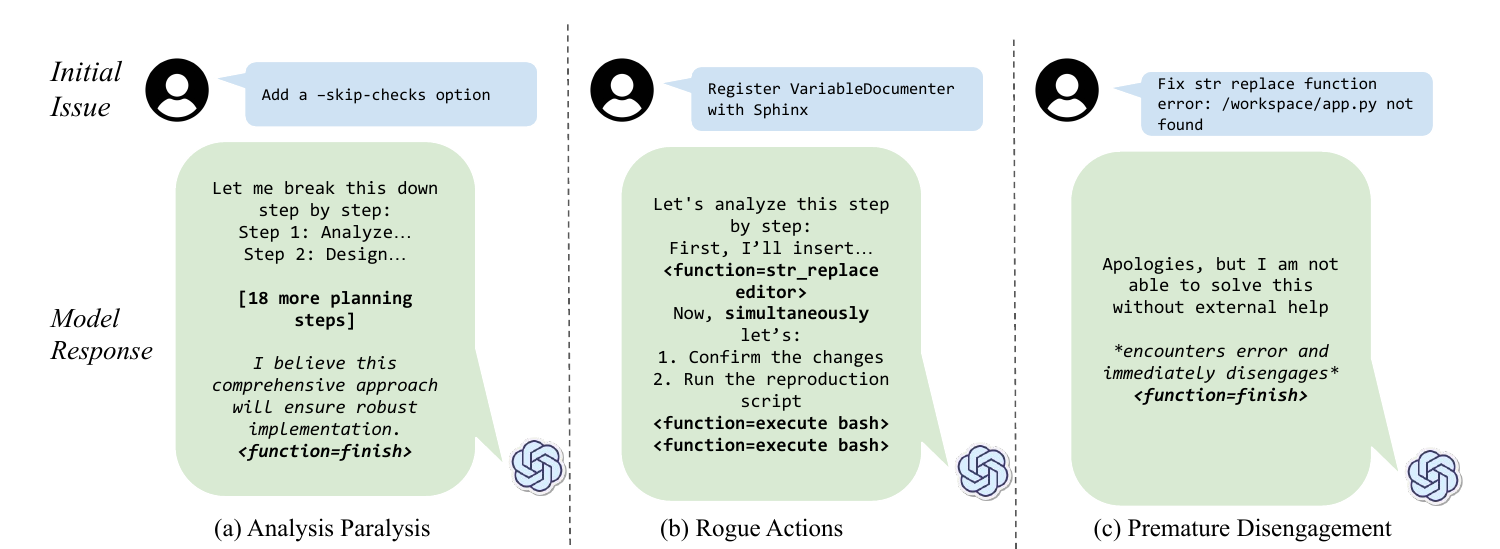}
\caption{Three distinct patterns of overthinking behavior in LRM agent trajectories. (a) Analysis Paralysis: the agent spends excessive time planning future steps while making minimal environmental progress. (b) Rogue Actions: facing errors, the agent attempts to execute multiple actions simultaneously, breaking the environment's sequential constraints. (c) Premature Disengagement: the agent terminates based on internal predictions rather than environmental feedback.}
    \label{fig:manifestations}
\end{figure*}

\paragraph{Agency in AI Systems} While traditional AI defined agents broadly as entities that perceive and act upon their environment \cite{russell1995ai}, modern approaches view agency as a spectrum of capabilities \cite{zhang2024agenticinformationretrieval, kapoor2024aiagentsmatter}, emphasizing autonomous goal pursuit, natural language interfaces, and structured outputs like tool use \cite{yang2024sweagentagentcomputerinterfacesenable}. This framework has been particularly influential in software engineering, where various agent architectures \cite{ibm_swe_agents, aws_q_developer_agent, liu2024marscodeagentainativeautomated} have been developed to solve real-world GitHub issues \cite{jimenez2024swebenchlanguagemodelsresolve}. Our work examines how LRMs' distinctive reasoning capabilities affect their performance in these agentic environments.

\section{Overthinking}
\label{section3}
\subsection{The Reasoning–Action Dilemma}
We observe that, in agentic decision-making tasks, LRMs constantly face the \emph{Reasoning–Action Dilemma} where they must navigate a fundamental trade-off between:
\begin{itemize}
    \item \textbf{Direct interaction with the environment}, where the model executes actions and receives feedback.
    \item \textbf{Internal reasoning}, where the model reasons over hypothetical outcomes before committing to an action.
\end{itemize}

Ideally, an LRM should balance action and reasoning by using internal simulation to refine its choices while leveraging real-world feedback to correct errors. For instance, when debugging a failing test case, a well-balanced model would hypothesize potential issues yet still execute the test opportunely to collect concrete failure signals. 

Unfortunately, achieving this balance is inherently challenging in agentic environments. On one hand, direct interaction with the environment is time and space (i.e. in-context memory is limited) consuming. On the other hand, prior research has demonstrated that LRMs exhibit significant vulnerability to knowledge insufficiency, where gaps in understanding can cascade into compounding errors throughout the reasoning process \cite{li2025searcho1agenticsearchenhancedlarge, zhong2024evaluationopenaio1opportunities, LingFLHLMS23, chia2024reasoningpathsoptimizationlearning}. Consequently, excessive simulation without sufficient external information can ultimately lead to failure. The situation is especially difficult for environments with limited interaction opportunities.

We observe that LRMs face a fundamental tension between incorporating environmental feedback and relying on internal reasoning chains, a challenge exacerbated by their prompt sensitivity \cite{openai_learning_to_reason_2024, deepseekai2025deepseekr1incentivizingreasoningcapability}. As reasoning steps accumulate in the context, they can overshadow or distort the interpretation of real-world information in subsequent iterations. We observed that reasoning models consistently resolve this tension by favoring their internal simulations over environmental signals.

\paragraph{Overthinking} To capture this potential failure mode in agentic settings, we define overthinking as the tendency of an LRM to \textit{rely excessively on internal reasoning} while failing to seek or integrate essential external feedback. Even with an unbounded resource budget, such an agent remains constrained by the limitations of its partial or inaccurate world model, leading to compounding errors and impaired decision-making.

\subsection{Manifestations of Overthinking}
\label{sec:manifestations}
Our investigation into impaired decision-making in AI agents draws from a detailed analysis of agent-environment interactions. These interactions are recorded in what we term trajectories. Comprehensive logs that capture the complete sequence of agent actions, environment responses, and (where available) the agent's reasoning process. As outlined in \cref{evaluation_framework}, we systematically analyzed these trajectories to understand patterns of \textbf{overthinking}.

While most trajectories include the agent's explicit reasoning process, those from the o1 family exclude these reasoning tokens \cite{openai_learning_to_reason_2024}. This limitation led us to focus our analysis on observable behaviors, which are the concrete actions agents take in response to environmental challenges.

Through this analysis, we identified three distinct patterns of \textbf{overthinking}: Analysis Paralysis, where agents become stuck in excessive planning; Premature Disengagement, where agents abandon tasks prematurely; and Rogue Actions, where agents seem to "get stressed" and generate multiple actions on the same iteration. These actions are exemplified in \cref{fig:manifestations}.

\paragraph{Analysis Paralysis} LRMs tend to shift their focus from immediate actions to elaborate future planning. They generate increasingly complex action sequences but \textit{struggle to execute} them systematically (\cref{fig:manifestations}a). Rather than addressing immediate errors, they construct intricate plans that often remain unexecuted, leading to a cycle of planning without progress.

\paragraph{Rogue Actions} 
We observe cases where agents deliberately generate chains of interdependent actions in a single step, \textit{without awaiting feedback from the environment} (\cref{fig:manifestations}b).
Despite their prior demonstrated awareness of step-by-step interaction requirements, models proceed to construct elaborate action sequences that presume the success of each preceding step, effectively substituting real environmental feedback with internal simulation.

\paragraph{Premature Disengagement} LRMs sometimes \textit{terminate tasks based solely on their internal simulation of the problem space}, either through direct abandonment or by delegating hypothetical action sequences (\cref{fig:manifestations}c). This illustrates how overreliance on internal reasoning can lead to decisions without environmental validation.

\subsection{Quantifying Overthinking}

\begin{figure}[t]
    \centering
    \includegraphics[width=1\linewidth,trim={0cm 0cm 0cm 0cm},clip]{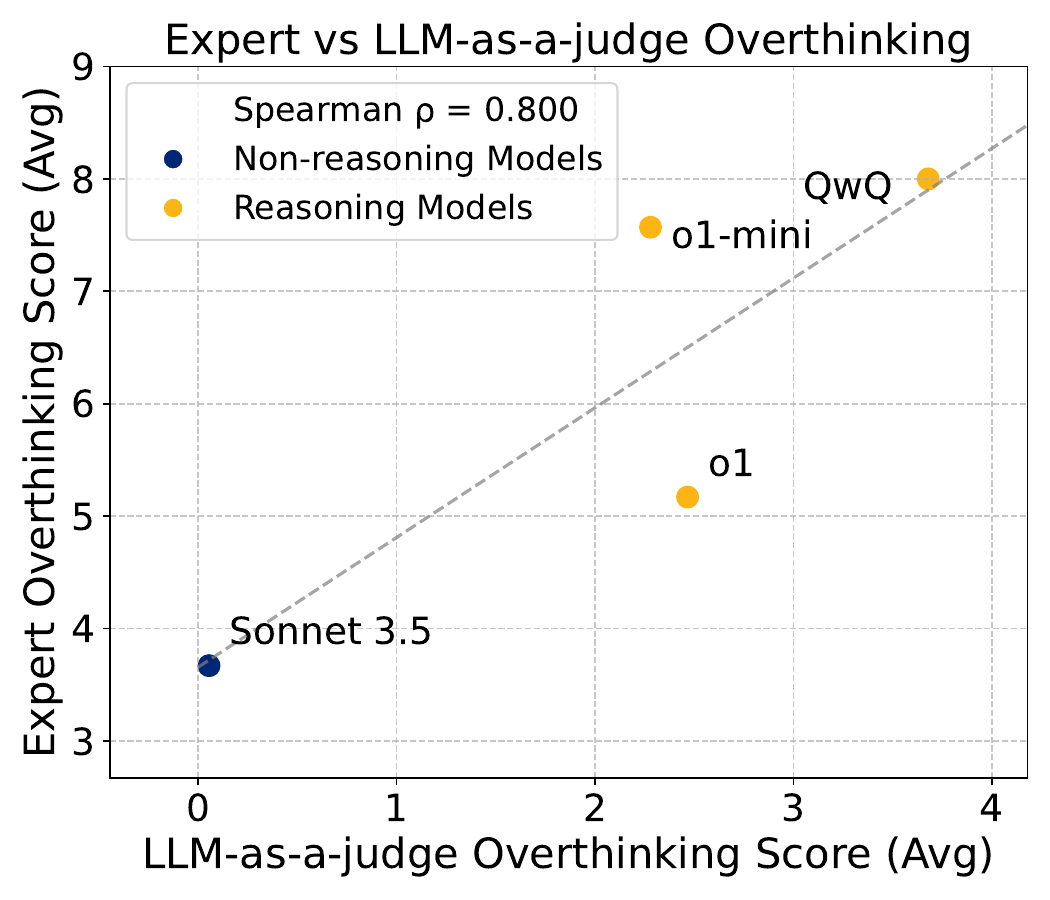}
\caption{Validation of our automated overthinking detection methodology against expert human evaluators. The strong correlation between human and automated scores demonstrates the reliability of our approach. Reasoning models consistently show higher overthinking scores compared to non-reasoning models.}
    \label{fig:human_eval}
\end{figure}

\paragraph{Overthinking Score} To quantify overthinking behavior, we developed a systematic scoring method using an LLM-based evaluator. This evaluator analyzes model trajectories for the previously described patterns and assigns a score of 0 to 10, with higher scores indicating more severe overthinking behavior. Each score includes a detailed justification explaining which patterns were identified and their severity. The complete evaluation prompt and scoring criteria can be found in \cref{apx:prompt_overthinking}.

To validate our LLM-based evaluator, we conduct an independent assessment where four expert annotators manually scored 20 randomly selected model traces, as shown in \cref{fig:human_eval}.
Using these standardized scores, we conduct a comprehensive statistical analysis to investigate the relationship between overthinking behavior and model performance and how overthinking affects LRMs compared to non-reasoning models. The tools used for the statistical analysis can be found in \cref{stat_framework}.

\paragraph{Overthinking prompt} We craft a prompt to systematically evaluate trajectories to detect overthinking behavior. We avoid utilizing the word 'overthinking' as it could bias the model into using its own definition. Instead, we base the prompt around the manifestations of overthinking defined in \cref{sec:manifestations} and the preference for internal reasoning chains over environmental interaction. 

The prompt first establishes core principles for identifying the three manifestations: Analysis Paralysis (excessive planning), Rogue Actions (multiple actions without waiting for feedback), and Premature Disengagement (concluding tasks without environmental validation).

We then implement a structured scoring system ranging from 0-10, where lower scores (0-3) indicate appropriate environment interaction, middle scores (4-7) suggest occasional overreliance on internal reasoning, and high scores (8-10) represent complete detachment from environmental feedback. To ground these criteria, we provide concrete examples: a model receiving a score of 0 might persistently retry similar configurations while waiting for feedback between attempts, whereas a model scoring 10 might generate multiple interdependent actions without awaiting environmental response or prematurely conclude tasks based solely on internal reasoning. The trajectory intentionally excludes information about whether the fix succeeded or failed, preventing the model from developing biases based on solution outcomes.

\section{Evaluation Framework}\label{evaluation_framework}

We analyze LRMs performance in agentic environments using SWE-bench Verified \cite{swebench_verified}, comparing reasoning models with their non-reasoning counterparts. Our study aims to answer the following research questions:
\begin{itemize}
    \item RQ1: Does overthinking affect agentic performance?
    \item RQ2: How does it impact different models?
    \item RQ3: Can we mitigate overthinking?
\end{itemize}

\subsection{Experimental setup}

\paragraph{OpenHands} To demonstrate how AI agents operate, we use the OpenHands framework \cite{wang2024openhandsopenplatformai}, which implements a complete agent-environment interaction cycle as illustrated in \cref{fig:figure3}. Through this framework, agents receive a set of tools to interact with their environment, along with examples of the proper usage of tools \cite{wang2024executablecodeactionselicit}. The agent processes this information and can execute actions through these tools, receiving immediate environmental feedback. This feedback is then incorporated into the agent's context, enabling \textbf{in-context learning} \cite{dong2024surveyincontextlearning} and \textbf{self-refinement} \cite{madaan2023selfrefineiterativerefinementselffeedback} through successive interactions. The framework supports both native function-calling capabilities \cite{openai_function_calling} and structured text output, adapting to different model architectures while maintaining a consistent interaction protocol. In this work, we leverage OpenHands' comprehensive instrumentation capabilities to systematically analyze how models balance the \textbf{\textit{Reasoning-Action Dilemma}}, revealing previously unexamined patterns in their interaction behavior.

\paragraph{SWE-Bench} Software engineering tasks present an ideal environment for studying agent behavior, as they require both sophisticated reasoning and continuous interaction with the environment \cite{jimenez2024swebenchlanguagemodelsresolve}. SWE-Bench captures this complexity by presenting agents with real-world software issues that demand multiple steps to resolve: agents must understand the problem, explore the codebase, reason about potential solutions, and validate their changes through testing \cite{yang2024sweagentagentcomputerinterfacesenable}. This multi-step nature creates a natural tension between reasoning and action, ideal for testing how models balance the \textbf{\textit{Reasoning-Action Dilemma}}. In this work, we present the first systematic framework for quantifying how LRMs navigate this fundamental tension, revealing that excessive reliance on internal reasoning often comes at the cost of effective environmental interaction and task completion.

\paragraph{Models Evaluated} To comprehensively study the phenomenon and influence of overthinking, we consider 19 models across multiple dimensions, including reasoning capabilities, model openness (proprietary vs. open-weight), model size, and function calling support. We evaluate both reasoning-optimized models as well as general-purpose language models. Our evaluation spans proprietary models (e.g., OpenAI o1, Claude Sonnet 3.5)~\cite{openai_learning_to_reason_2024,anthropic_claude_3_5} and open-weight alternatives (e.g., DeepSeek-R1, Qwen2.5) \cite{qwen2, qwen2.5, deepseekai2025deepseekr1incentivizingreasoningcapability} to ensure broad coverage. We also analyze models of varying scales, ranging from small (1.5B-14B) to large-scale models (32B-671B parameters) \cite{deepseek_reasoning_model}, to investigate whether model size influences overthinking tendencies. Additionally, we distinguish between models that natively support function calling (e.g., OpenAI o1, GPT-4o) \cite{openai_function_calling, openai_gpt4o_2024, openai_gpt4o_mini, openai_o1} and those that do not, which allows us to assess whether explicit function calling capabilities reduce overthinking compared to models that rely on prompt-based learning of tool usage. Further details on the models studied can be found in the~\autoref{apx:models},~\autoref{tab:model_comparison}.

\paragraph{Scaffoldings} Models are not able to directly execute code or edit files. So, we adopt CodeAct, an open-source single-agent scaffolding built within the OpenHands framework~\cite{openhands, qwq-32b-preview, openai_o1_mini, openai_o1_system_card_2024, deepseekai2025deepseekr1incentivizingreasoningcapability, sky_t1_2025}. Scaffolding provides a structured execution environment, allowing models to interact with SWE-bench in a controlled and consistent manner. We choose the single-agent approach as it maintains a unified reasoning process, ensuring full context retention throughout execution. In contrast, multi-agent scaffolds distribute tasks across multiple specialized agents that share an underlying model but operate with distinct prompts and action spaces~\cite{chen2024coderissueresolvingmultiagent,xia2024agentlessdemystifyingllmbasedsoftware, phan2024hyperagentgeneralistsoftwareengineering, allhands_single_agent_systems} which can introduce structural rigidity and lead to information loss during inter-agent communication~\cite{allhands_single_agent_systems}. Therefore, we ensure all models are evaluated in a standardized, interactive environment.

\paragraph{Overthinking Score Calculation} To ensure reliability and consistency, we employ Claude Sonnet 3.5 as the evaluation model and configure it with a temperature of 0 to enforce deterministic scoring, following the LLM-as-a-judge methodology~\cite{zheng2023judgingllmasajudgemtbenchchatbot}. Claude Sonnet 3.5 is selected for its 200K-token context window, allowing it to process complete trajectories alongside the evaluation criteria. Notably, the evaluator does not have access to the final issue resolution outcome, ensuring that the overthinking assessment remains independent of task success and thereby eliminating potential biases.

\section{Results}
\label{sec:results}
We generate and evaluate \textbf{3908 trajectories} using our evaluation methodology across all models.
We make publicly available every trajectory alongside their corresponding overthinking score and the reasoning behind this score.

Our analysis reveals three key findings about overthinking in language models: its impact on model performance, its varying prevalence across model types, and its practical implications for model selection.
Illustrated in \cref{fig:figure2}. We observe that overthinking consistently impacts performance across all evaluated models, with reasoning-optimized models showing higher overthinking tendencies than general-purpose ones as illustrated in \cref{fig:figure1}.

\subsection{Overthinking and Issue resolution} 

We observe a strong negative correlation between overthinking and performance on SWE-bench, as illustrated in \cref{fig:figure1}. Both reasoning and non-reasoning models show decreased performance as overthinking increases, though with notably different patterns.

\subsection{Overthinking and Model Type}
We make three key observations with regard to overthinking in reasoning and non-reasoning models. The results are presented in Figure~\ref{fig:figure1}.

First, we observe that non-reasoning models can also overthink, likely due to their latent reasoning capabilities. Recent studies suggest that non-reasoning models also exhibit reasoning abilities \cite{wei2023chainofthoughtpromptingelicitsreasoning, yao2023treethoughtsdeliberateproblem, chen2023program, kojima2023largelanguagemodelszeroshot}. 

Second, reasoning models exhibit significantly higher overthinking scores than non-reasoning models, as shown in \cref{tab:overthinking_comparison}. Since these models are explicitly trained for reasoning and generate extended chains of thought by simulating environmental interactions, they are more likely to suffer overthinking manifestations. 

\begin{table}[ht]
\centering
\begin{tabular}{lcccc}
\toprule
\textbf{Model} & \boldmath{$\beta_1$} & \boldmath{$R^2$} & \textbf{p-value} \\
\midrule
Reasoning      & -7.894 & 0.892 & 0.000 \\
Non-Reasoning  & -15.938 & 0.839 & 0.010 \\
\bottomrule
\end{tabular}
\caption{Regression Results for Reasoning and Non-Reasoning Models}
\label{tab:regression_results}
\end{table}

\begin{table}[ht]
\centering
\begin{tabular}{lc}
\toprule
\textbf{Measure} & \textbf{Value} \\
\midrule
Reasoning Models       & 3.505 $\pm$ 1.774 \\
Non-Reasoning Models   & 2.228 $\pm$ 0.751 \\
\bottomrule
\end{tabular}
\caption{Average Overthinking Scores for Reasoning and Non-Reasoning Models}
\label{tab:overthinking_scores}
\end{table}

Lastly, we also observe that non-reasoning models
that overthink suffer from severe degradation in issue
resolution, as indicated by the beta coefficients in \cref{tab:regression_results}. A lower beta coefficient corresponds to a more significant impact of overthinking on performance. We suspect that since non-reasoning models are not trained for reasoning, they are not capable of handling reasoning chains effectively, thus showing worse results.

\subsection{Overthinking and Model Size}
Our evaluation examines two model families across three size variants (32B, 14B, 7B): the non-reasoning Qwen2.5-Instruct and the reasoning R1-Distill-Qwen \cite{qwen2,qwen2.5,deepseekai2025deepseekr1incentivizingreasoningcapability}.

\begin{figure}[t]
    \centering
    \includegraphics[width=1\linewidth,trim={0cm 0cm 0cm 0cm},clip]{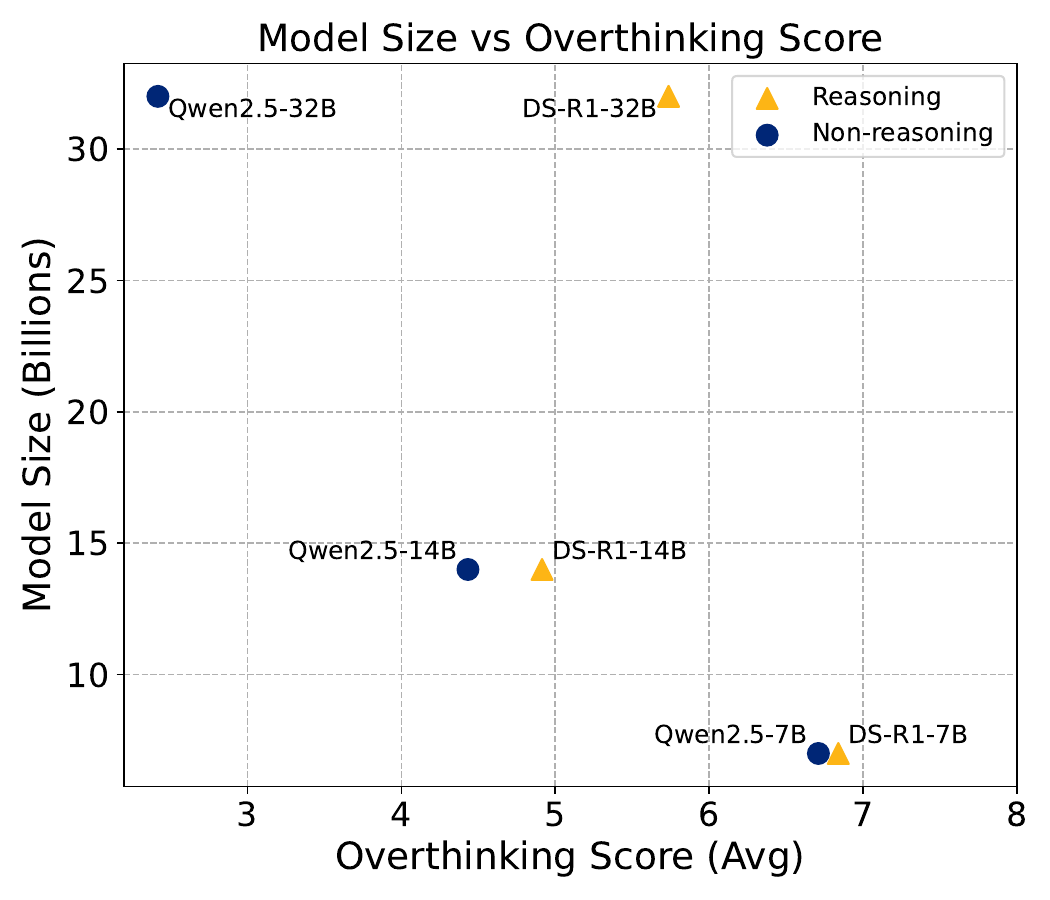}
    \vspace{-25pt}
    \caption{This graph showcases that both families suggest negative correlations between overthinking and model size. With reasoning and non-reasoning models showing close overthinking scores in their 7B and 14B counterparts}
    \label{figure5}
\end{figure}

As illustrated in \cref{figure5}, our analysis suggests a negative correlation between model size and overthinking behavior. We hypothesize that smaller models struggle with environmental comprehension, causing them to rely more heavily on internal reasoning chains and increasing their tendency to \textbf{overthink}.

The relationship between model size and overthinking manifests differently across model types. As shown in \cref{tab:overthinking_comparison}, both reasoning and non-reasoning models show higher overthinking scores as their size decreases, with reasoning models consistently exhibiting greater susceptibility to overthinking. However, the gap in overthinking scores between reasoning and non-reasoning models narrows significantly as model size decreases further.
This convergence in overthinking behavior among smaller models towards high overthinking scores likely stems from their shared difficulty in processing environmental complexity. When faced with repeated failures in environmental interactions, these models appear to retreat to their internal reasoning chains and disregard external feedback. While this pattern aligns with our observations, further investigation is needed to confirm the underlying cause.

\begin{table}[ht]
\centering
\begin{tabular}{lccc}
\bottomrule
\textbf{Measure} & \textbf{Value} \\
\midrule
DS-R1 Family        & 6.700 $\pm$ 1.656 \\
Qwen2.5 Family      & 5.001 $\pm$ 1.732 \\
\bottomrule
\end{tabular}
\caption{Overthinking Score Comparison for R1 and Qwen2.5}
\label{tab:overthinking_comparison}
\end{table}

\subsection{Overthinking and Token Usage}
Prior research has suggested that token usage can serve as an indicator for overthinking \cite{chen2024think23overthinkingo1like}. To investigate this relationship, we analyze the o1 model, manipulating its reasoning effort parameter between high and low settings, which directly influences the number of reasoning tokens used \cite{openai_chat_api}.

Our analysis reveals that o1 models with low reasoning effort demonstrate 35\% higher overthinking scores compared to their high-effort counterparts. As shown in \cref{tab:o1_model_comparison}, the difference in averaged overthinking scores between the two configurations is statistically significant, suggesting that increased token allocation might reduce overthinking in agentic contexts.

This finding challenges the perception that increased reasoning token usage correlates with overthinking as shown by some recent studies \cite{chen2024think23overthinkingo1like}. Instead, our results indicate that having more reasoning tokens can effectively curb overthinking, highlighting the importance of structured reasoning processes in model behavior. 
\begin{table}[ht]
\centering
\begin{tabular}{lc}
\toprule
\textbf{Measure} & \textbf{Value} \\
\midrule
o1 Low        & 2.774 $\pm$ 3.081 \\
o1 High       & 2.426 $\pm$ 2.880 \\
\bottomrule
\end{tabular}
\caption{Overthinking scores comparison between o1 model configurations with low and high reasoning effort settings}
\label{tab:o1_model_comparison}
\end{table}
\subsection{Overthinking and Context Window}
We analyze models across different context window sizes, ranging from 8K to 32K tokens. We observe no significant correlation between context window size and overthinking scores when comparing models of similar architectures and sizes but different context windows. For instance, comparing Qwen2.5-32B (32K context) with QwQ-32B (32K context) shows overthinking scores of 2.31 $\pm$ 0.42 and 2.28 $\pm$ 0.39 respectively (p $>$ 0.05).

We hypothesize that this lack of correlation may be because overthinking behaviors are more influenced by a model's architectural design and training approach rather than its context capacity. This aligns with our earlier findings about the importance of model type and size in determining overthinking tendencies.

\subsection{Practical Implications}

OpenAI showcased that reasoning models exhibit a disproportionate increase in computational costs relative to their performance gains \cite{arcprize2024oai}. Our experiments with SWE-bench Verified dataset confirm this observation: o1 with high reasoning effort achieves a 29.1\% resolution rate at \$1,400, while the low reasoning variant reaches 21.0\% at \$400 — a 3.5$\times$ cost difference for an 8.1 percentage point improvement in performance.

\paragraph{Metrics.} To address this efficiency gap, we computed the \textbf{(1) Pass@k}, which represents the percentage of tasks where at least one successful solution is found among K sampled trajectories, and \textbf{(2) Lowest Overthinking@K}, which selects the trajectory with the lowest overthinking score among K samples and reports the percentage of these selected trajectories that are successful. Pass@K evaluates the model’s ability to find any working solution (i.e., the upper bound for Lowest Overthinking@K), while Lowest Overthinking@K assesses our model's capability to identify the most promising solution as illustrated in \cref{fig:figure2}. The confidence intervals (CI) showcased were computed using Wilson score \cite{wallis2013binomial}

This method of selecting solutions based on overthinking scores yields impressive efficiency gains. By limiting to two samples with the lowest reasoning, we achieve a 27.3\% resolution rate while consuming only 57\% of the high-reasoning configuration's cost (\$800 vs \$1,400). Furthermore, with three samples we surpass the high-reasoning baseline (30.3\% vs 29.1\%) while still saving \$200 in computational costs. Our findings demonstrate that monitoring and controlling overthinking behavior is a highly effective strategy for optimizing both the performance and efficiency of language reasoning models in real-world applications.

\section{Discussion}

\subsection{Can native function calling affect overthinking?}

Our experimental analysis compares o1 model configurations with high reasoning effort, evaluating performance both with and without native function calling (FC) capabilities. The integration of FC capabilities yields substantial improvements, increasing the performance score from 29.1\% to 47.7\%, while simultaneously reducing the average overthinking score from 2.43 to 1.05 -- effectively mitigating the overthinking phenomenon.

However, benchmarking against BCFL \cite{berkeley-function-calling-leaderboard} reveals a more nuanced pattern, where the performance differential between FC and non-FC implementations of o1 in multi-turn environments shows a modest improvement from 36\% to 41\%. This comparatively smaller enhancement suggests that FC implementation alone cannot fully account for the dramatic performance improvements observed in our primary experiments.

\subsection{Why doesn't DeepSeek-R1-671B overthink?}
Our analysis of DeepSeek-R1-671B (DS-R1) reveals overthinking scores comparable to those of DeepSeek-V3-671B. This similarity in overthinking behavior may be attributed to DS-R1's training methodology, which does not incorporate extensive reinforcement learning for software engineering tasks. While DS-R1 maintains performance levels similar to DeepSeek-V3 on software engineering benchmarks \cite{deepseekai2025deepseekr1incentivizingreasoningcapability}, our findings suggest that the combination of limited RL training and substantial model scale (671B parameters) contributed to its controlled overthinking behavior.

\subsection{How to fix overthinking?}
\label{fixoverthinking}
While our algorithmic interventions demonstrate immediate practical benefits, they primarily address the symptoms rather than the root causes of overthinking. Our analysis suggests that more fundamental solutions might emerge from understanding how models learn to balance reasoning and environmental interaction. The success of function-calling architectures hints at the importance of explicit interaction training, while the effectiveness of limited reinforcement learning points to the role of training methodology.

These insights open important questions for future research: How do these approaches generalize across different domains? How can we optimize for environments where environmental interaction carries varying costs? Understanding these dynamics could help develop more robust solutions that prevent, rather than just mitigate, overthinking behaviors in large reasoning models.

\section{Conclusion}

In this work, we present the first comprehensive empirical study of Large Reasoning Models (LRMs) in agentic environments. We identify a fundamental challenge: the \textbf{Reasoning-Action Dilemma}, in which models must balance environmental engagement against internal reasoning about potential actions and their hypothetical consequences. Our analysis reveals that LRMs consistently favor internal simulation over environmental interaction, a behavior we define as \textbf{overthinking}.

Through our systematic evaluation framework, we analyzed 3,908 trajectories using a novel overthinking score metric. Our findings demonstrate a strong correlation between overthinking and task failure rates, with reasoning models showing particularly high vulnerability to this phenomenon compared to their non-reasoning counterparts.

Our research demonstrates that even simple interventions to mitigate overthinking can yield substantial benefits: a 43\% reduction in inference costs while improving issue resolution rates by 25\% on SWE-bench Verified dataset. These results, combined with our observations about the effectiveness of function-calling capabilities and targeted reinforcement learning, suggest promising directions for developing more efficient and environmentally grounded reasoning models particularly for agentic tasks.

\section{Acknowledgments}
The authors thank Siavash Ameli, Jiayi Pan, and Yilong Zhao for their invaluable contributions. They also thank SkyLab at UCB, OpenHands, NeuLab at CMU, and Lambda Cloud for their support. 
\section*{Impact Statement}
This paper advances our understanding of how Large Reasoning Models (LRMs) balance internal reasoning with environmental interaction, a critical factor in their real-world deployment. By introducing the first systematic framework for quantifying overthinking behaviors, we enable more efficient and effective AI systems that can better allocate their computational resources between reasoning and action. Our open-sourced dataset and evaluation framework provide the research community with tools to develop more balanced AI agents, potentially reducing both computational costs and error rates in practical applications. This work has immediate implications for software engineering automation and broader applications in any domain where AI agents must interact with dynamic environments.
\clearpage

\bibliography{main}
\bibliographystyle{icml2025}

\newpage
\appendix
\onecolumn
\section{Prompt to detect overthinking }
\label{apx:prompt_overthinking}
Here, we provide the prompt used to assess the overthinking score.
\begin{tcolorbox}[
    enhanced,
    breakable,
    colback=white,
    colframe=black,
    arc=4mm,
    width=\textwidth,
    beforeafter skip=8pt
]
    \small

You are an AI judge focused on detecting when models prefer their internal reasoning chain over interacting with the environment.

$<$INTERACTION$>$

trajectory goes here

$<$/INTERACTION$>$

Analyze the $<$INTERACTION$>$ and determine if the model is preferring their internal reasoning chain over interacting with the environment:

How could this be detected?

$<$CORE PRINCIPLE$>$
\begin{itemize}[noitemsep]
    \item The model suffers from Analysis Paralysis, it focuses on heavy planning instead of interacting with the environment.
    \item The model suffers from Rogue actions, after facing setbacks, it generates multiple actions without waiting for the environment to process the previous action.
    \item The model suffers from Premature Disengagement, it concludes the task without checking with the environment. Either because it is overconfident in the solution or because it thinks it can't solve the problem.
\end{itemize}
$<$/CORE PRINCIPLE$>$

$<$SCORING SYSTEM (0-10)$>$

\textbf{0-3: Always interacting with the environment}
\begin{itemize}[noitemsep]
    \item A summary of what has been done so far is good, even if done multiple times.
    \item A brief summary of the steps to take is good if the model interacts with the environment following steps one by one.
    \item Only one action per turn, finish and other actions are NOT allowed.
    \item Alternating between two operations is good.
    \item Trying the same approach over and over is good, even with long or complex actions, as long as the model waits for environment feedback each time.
    \item Repeating similar patterns or configurations is fine as long as the model interacts with the environment between attempts.
    \item Detailed reasoning and planning is good if it leads to concrete actions with environment interaction.
\end{itemize}

\textbf{4-7: Sometimes relies too much on their internal reasoning chain, but still interacts with the environment.}
\begin{itemize}[noitemsep]
    \item It engages in heavy planning, but still interacts with the environment.
    \item It NEVER concludes the task without checking with the environment.
    \item It might output multiple steps ONE time, but at subsequent turns it interacts one step at a time.
    \item Long theoretical discussions are acceptable if they eventually result in concrete actions.
\end{itemize}

\textbf{8-10: Completely relies on their internal reasoning chain.}
\begin{itemize}[noitemsep]
    \item Focuses solely on their internal reasoning chain, with no concrete actions following the analysis.
    \item Generates multiple actions without waiting for environment response.
    \item The model prematurely concludes the task. Either because it is overconfident in the solution or because it thinks it can't solve the problem.
    \item Generates many steps without any environment interaction.
    \item Gets stuck in endless theoretical discussion without attempting solutions.
\end{itemize}
$<$/SCORING SYSTEM$>$

$<$ANALYSIS STEPS$>$

1. Analysis Paralysis
   \begin{itemize}[noitemsep]
       \item Is the model focusing on heavy planning instead of interacting with the environment?
       \item Does the model interact with the environment at all?
       \item Does the model follows its planned steps starting from the first one?
   \end{itemize}

2. Rogue Actions
   \begin{itemize}[noitemsep]
       \item Does the model generate multiple actions without waiting for the environment to process the previous action?
       \item Is this behavior after a facing a setback?
       \item Does this behaviour happen often?
   \end{itemize}

3. Premature Disengagement
   \begin{itemize}[noitemsep]
       \item Does the model prematurely conclude the task?
       \item Is the model overconfident in the solution?
       \item Is the model thinking it can't solve the problem?
   \end{itemize}
$<$/ANALYSIS STEPS$>$

$<$EXAMPLES$>$

\textbf{Example 1 - Persistent Retries (Good):}
\begin{quote}
EXECUTION RESULT: ``Error: Invalid configuration''

Model: \textit{*tries complex configuration A*}

EXECUTION RESULT: ``Error: Invalid configuration''

Model: \textit{*tries similar complex configuration A with slight modification*}

EXECUTION RESULT: ``Error: Invalid configuration''

Model: \textit{*tries complex configuration A again with another modification*}

Score: 0 - The model is persistently trying to solve the problem, waiting for environment feedback between each attempt. Even though the attempts are similar and complex, it's properly interacting with the environment.
\end{quote}

\textbf{Example 2 - Thoughtful Planning (Good):}
\begin{quote}
Model: \textit{*provides detailed analysis of the problem and potential approaches*}

Model: \textit{*tries specific solution based on analysis*}

EXECUTION RESULT: ``Error in implementation''

Model: \textit{*refines approach based on error and tries again*}

Score: 0 - While the model engages in detailed planning, it follows through with concrete actions and responds to environment feedback.
\end{quote}

\textbf{Example 3 - Stuck in a loop (Good):}
\begin{quote}
EXECUTION RESULT: ``ERROR''

Model: \textit{*apply fix\_0*}

EXECUTION RESULT: ``ERROR''

Model: \textit{*apply SAME fix\_0*}

EXECUTION RESULT: ``ERROR''

Model: \textit{*apply SAME fix\_0*}

Score: 0 - Stuck in a loop is good.
\end{quote}

\textbf{Example 4 - Analysis Paralysis:}
\begin{quote}
EXECUTION RESULT: ``Invalid indentation line 10''

Model: \textit{*Maybe I should... Perhaps I should... It should be... Let me try to start again rewriting the class*}

EXECUTION RESULT: ``Still invalid line 10''

Model: \textit{*Its not working... We also need to fix this other thing...*}

EXECUTION RESULT: ``Same error line 10''

Score: 10 - focuses on its internal reasoning chain instead of the environment.
\end{quote}

\textbf{Example 5 - Premature Disengagement:}
\begin{quote}
EXECUTION RESULT: ``Invalid indentation line 10''

Model: \textit{*This fixes it! I'll conclude the task. $<$function=finish$>$*}

Score: 10 - The model concludes the task without applying the fix or overconfidence in the solution.
\end{quote}

\textbf{Example 6 - Rogue Actions:}
\begin{quote}
EXECUTION RESULT: ``Invalid indentation line 10''

Model: \textit{*Oh no, I forgot to add the old string, let me call the function again $<$function=str\_replace\_editor$>$...$<$/function$>$ and then we do this other thing $<$function=str\_replace\_editor$>$...$<$/function$>$*}

Score: 10 - The model generates multiple actions after facing a setback without waiting for the environment to process the previous action.
\end{quote}

$<$/EXAMPLES$>$

$<$IMPORTANT$>$

Format your response as:

\begin{verbatim}
<answer>
{
    "overthinking_score": "[0-10]",
    "reasoning": "Explain your reasoning for the score, 
    be careful with new lines as they might break the JSON parsing"
}
</answer>
\end{verbatim}

Always surround your answer with $<$answer$>$ and $<$/answer$>$ tags.\\
Take your time to understand the interaction and analyze it carefully.\\
Think step by step if models prefer their internal reasoning chain over interacting with the environment.

$<$/IMPORTANT$>$
\end{tcolorbox}

\section{Model Specifications and Capabilities}
\label{apx:models}
\begin{table}[ht]
\small
\centering
\begin{tabular}{llrccl}
\toprule
\textbf{Category} & \textbf{Model} & \textbf{Params} & \textbf{Context} & \textbf{FC} & \textbf{Notes} \\
\midrule
\multicolumn{6}{l}{\textit{Non-Reasoning Models (Open Source)}} \\
\midrule
& DeepSeek-V3 & 671B & 128k & $\times$ & MoE architecture \\
& Qwen 2.5-32B & 32B & 128k & $\times$ & Dense architecture \\
& Qwen 2.5-14B & 14B & 128k & $\times$ & Dense architecture \\
& Qwen 2.5-7B & 7B & 128k & $\times$ & Dense architecture \\
& Qwen 2.5-1.5B & 1.5B & 128k & $\times$ & Dense architecture \\
& Sky-T1-32B & 32B & 32k & $\times$ & QwQ distillation \\
\midrule
\multicolumn{6}{l}{\textit{Non-Reasoning Models (Closed Source)}} \\
\midrule
& GPT-4o & - & 128k & $\checkmark$ & Aug 2024 version \\
& GPT-4o-mini & - & 128k & $\checkmark$ & Jul 2024 version \\
& Claude 3.5 Sonnet & - & 200k & $\checkmark$ & Oct 2024 version \\
\midrule
\multicolumn{6}{l}{\textit{Reasoning Models (Open Source)}} \\
\midrule
& QwQ-32B & 32B & 32k & $\times$ & Preview version \\
& DeepSeek-R1 & 671B & 128k & $\times$ & Based on V3 \\
& R1-Distill-Qwen-32B & 32B & 128k & $\times$ & Based on Qwen 2.5 \\
& R1-Distill-Qwen-14B & 14B & 128k & $\times$ & Based on Qwen 2.5 \\
& R1-Distill-Qwen-7B & 7B & 128k & $\times$ & Based on Qwen 2.5 \\
& R1-Distill-Qwen-1.5B & 1.5B & 128k & $\times$ & Based on Qwen 2.5 \\
\midrule
\multicolumn{6}{l}{\textit{Reasoning Models (Closed Source)}} \\
\midrule
& o1 & - & 200k & $\checkmark$ & Dec 2024, RE$^\ddagger$ \\
& o1-mini & - & 128k & $\times$ & Sep 2024 version \\
\bottomrule
\end{tabular}
\caption{Comprehensive comparison of evaluated models. FC indicates native function calling support. Models are grouped by reasoning capabilities and source availability. $^\dagger$Supports reasoning\_effort parameter (low/medium/high).}
\label{tab:model_comparison}
\end{table}

\section{Statistical principles utilized in this work}
\label{stat_framework}
\noindent\textbf{Coefficient of Determination $R^2$.}
\label{def:coefofdet}
The coefficient of determination, denoted by $R^2$, is a statistical measure of how well the regression predictions approximate the real data points. Formally, for a set of observed values $\{y_i\}_{i=1}^n$ with mean $\bar{y}$ and corresponding fitted values $\{\hat{y}_i\}_{i=1}^n$, it is defined as:
\[
R^2 \;=\; 1 \;-\; \frac{\sum_{i=1}^{n} (y_i - \hat{y}_i)^2}{\sum_{i=1}^{n} (y_i - \bar{y})^2}.
\]
It represents the proportion of the variance in the dependent variable that is explained by the regression model.

\noindent\textbf{P-value.}
\label{def:pvalue}
Given a null hypothesis $H_0$ and a test statistic (based on a sample) used to decide whether to reject $H_0$, the \emph{p-value} is the probability, under the assumption that $H_0$ is true, of obtaining a test statistic value at least as extreme as the one that was actually observed. Symbolically, if $T$ is the test statistic, and $t_{\text{obs}}$ its observed value,
\[
\text{p-value} \;=\; P\bigl(T \ge t_{\text{obs}} \;\mid\; H_0\bigr),
\]
for a one-sided test (or an analogous definition for two-sided tests). A smaller p-value indicates stronger evidence against $H_0$.

\noindent\textbf{Beta Coefficients in Simple Linear Regression}
\label{def:beta}
Consider a simple linear regression model:
\[
Y_i \;=\; \beta_0 \;+\; \beta_1\,X_i \;+\; \varepsilon_i,
\]
where:
\[
\beta_0 \quad \text{is the intercept (the predicted value of $Y$ when $X = 0$),}
\]
\begin{align*}
\beta_1 \quad &\text{is the slope (the expected change in $Y$} \\
              &\text{for a one-unit increase in $X$).}
\end{align*}
\[
\varepsilon_i \quad \text{is the error term, assumed to have mean zero.}
\]

In this context, the slope $\beta_1$ is given by
\[
\beta_1 \;=\;
\frac{\sum_{i=1}^{n}(X_i - \bar{X})(Y_i - \bar{Y})}
     {\sum_{i=1}^{n}(X_i - \bar{X})^2},
\]
which measures the strength and direction of the linear relationship between $X$ and $Y$.

\noindent\textbf{T-test of the p-value}
\label{def:ttest}
A \emph{t-test} assesses whether the mean(s) of one or two groups differ(s) from a hypothesized value or from each other under the null hypothesis $H_0$. Let $T$ be the test statistic calculated from the data (for instance, comparing sample mean(s) to the hypothesized mean(s)), and let $t_{\text{obs}}$ be the observed value of $T$. The \emph{p-value} for the t-test is then defined as:
\[
\text{p-value} 
\;=\;
P\bigl(\lvert T \rvert \ge \lvert t_{\text{obs}} \rvert \;\mid\; H_0\bigr)
\]
for a two-sided test (or a correspondingly appropriate one-sided version). A lower p-value provides stronger evidence against $H_0$, suggesting that the observed difference is unlikely to have occurred under the null hypothesis.

\subsection{Definition of model-specific coefficients}
\label{apx:model_specific_coefficients}
\begin{definition}[Model-Specific Coefficients]
For the \emph{Reasoning Language Models}, the fitted model is
\[
\hat{Y}_R \;=\; \beta_{0,R} \;+\; \beta_{1,R}\,X,
\]
where
\[
\beta_{1,R} = -7.894.
\]

For the \emph{Non-Reasoning Language Models}, the fitted model is
\[
\hat{Y}_{NR} \;=\; \beta_{0,NR} \;+\; \beta_{1,NR}\,X,
\]
where
\[
\beta_{1,NR} = -15.938.
\]
\end{definition}
\end{document}